\documentclass[preprint,12pt]{elsarticle}

\usepackage{amssymb}
\usepackage{subfigure}
\usepackage{setspace}
\usepackage{geometry} 
\usepackage{hyperref}
\usepackage{multirow}
\usepackage{upgreek}
\usepackage{amsmath}
\usepackage{mathrsfs} 
\usepackage{algorithm}
\usepackage{algpseudocode}
\usepackage[T1]{fontenc} 
\usepackage{kpfonts} 
\usepackage{xcolor}

\usepackage{natbib}

\usepackage{lineno}

\journal{Information and Management}

\begin{document}

\begin{frontmatter}




\title{Unveiling the Potential of Knowledge-Prompted ChatGPT for Enhancing Drug Trafficking Detection on Social Media}
\author[inst1]{Chuanbo Hu}

\affiliation[inst1]{organization={Lane Department of Computer Science and Electrical Engineering, West Virginia University},
            addressline={395 Evansdale Dr}, 
            city={Morgantown},
            postcode={26505}, 
            state={West Virginia},
            country={United States}}
            
\author[inst2]{Bin Liu}
\affiliation[inst2]{organization={Department of Management Information Systems, West Virginia University},
            addressline={83 Beechurst Avenue}, 
            city={Morgantown},
            postcode={26505}, 
            state={West Virginia},
            country={United States}}

\author[inst1]{Xin Li}

\author[inst3]{Yanfang Ye}

\affiliation[inst3]{organization={Department of Computer Science and Engineering, University of Notre Dame},
            addressline={257 Fitzpatrick Hall of Engineering}, 
            city={Notre Dame},
            postcode={46556}, 
            state={IN},
            country={United States}}

\begin{abstract}
Social media platforms such as Instagram and Twitter have emerged as critical channels for drug marketing and illegal sale. Detecting and labeling online illicit drug trafficking activities becomes important in addressing this issue. However, the effectiveness of  conventional supervised learning methods in detecting drug trafficking heavily relies on having access to substantial amounts of labeled data, while data annotation is time-consuming and resource-intensive. Furthermore, these models often face challenges in accurately identifying trafficking activities when drug dealers  use deceptive language and euphemisms to avoid detection. 
To overcome this limitation, we conduct the first systematic study on leveraging large language models (LLMs), such as ChatGPT,  to detect illicit drug trafficking activities on social media. We propose an analytical framework to compose \emph{knowledge-informed prompts}, which serve as the interface that humans can interact with and use LLMs to perform the detection task.
Additionally, we design a Monte Carlo dropout based prompt optimization method to further to improve performance and interpretability. 
Our experimental findings demonstrate that the proposed framework outperforms other baseline language models in terms of drug trafficking detection accuracy, showing a remarkable improvement of nearly 12\%. By integrating prior knowledge and the proposed prompts, ChatGPT can effectively identify and label drug trafficking activities on social networks, even in the presence of deceptive language and euphemisms used by drug dealers to evade detection. The implications of our research extend to social networks, emphasizing the importance of incorporating prior knowledge and scenario-based prompts into analytical tools to improve online security and public safety.
\end{abstract}

\begin{highlights}
\item We conduct the first systematic study on leveraging large language models (LLMs), such as ChatGPT,  to detect illicit drug trafficking activities on social media. 
\item We propose an analytical framework to compose knowledge-informed prompts, which serve as the interface that humans can interact with and use LLMs for the detection of drug trafficking activities.
\item We design a prompt optimization method to further enhance  the effectiveness of the prompts. 
\item We demonstrate the effectiveness of the proposed framework through extensive experiments on an illicit drug  trafficking dateset we collected from Instagram. 
\end{highlights}


\end{frontmatter}


\section{Introduction}
\label{sec:sample1}
Drug trafficking, the illegal sale or transport of prohibited drugs,  is a global issue that has far-reaching impacts on communities, families, and individuals. Illegal drug trade and usage lead to addiction and health problems and have broader social impacts. Drug trafficking organizations are often associated with violence, corruption, and other forms of criminal activity \cite{morris2012corruption,morris2013drug,doherty2007uav}. 
Social media has provided drug dealers with a convenient platform to market and sell their illicit products \cite{demant2019drug,liang2011prevalence}. Social media platforms offer a broad audience reach and provide drug dealers with a level of anonymity that was previously unattainable. These platforms also facilitate communication between dealers and customers, enabling them to coordinate transactions \cite{moyle2019drugsforsale}. 
Detection of drug trafficking on social networks has become critical in tackling drug trafficking. However, detecting drug trafficking activities on social media poses difficulties due to the use of disguised language and euphemisms by drug dealers \cite{hu2021detection}. Drug dealers employ code words, acronyms, and other disguised language to avoid detection by law enforcement agencies and social media platforms. Consequently, although social media platforms have implemented some mechanisms to combat drug trafficking, they are ineffective due to the aforementioned challenges.


Several research studies are building task-specific machine learning models to detect drug trafficking activities on social media, including detection of drug dealers \cite{hu2021identifying,li2019machine}, drug trafficking events \cite{yang2017tracking, hu2021detection,ZhaoTOCBB22}, and drug-related hashtags \cite{hu2023fine}. 
The problem of detection of illicit drug trafficking can be formulated as a supervised machine learning problem, where the input is typically the text or/and images on social media, and the output is drug trafficking activities to be detected. As such, different machine learning approaches,  such as Convolutional Neural Networks (CNN) \cite{hu2021detection,hu2021identifying,ZhaoTOCBB22}, long short-term memory networks (LSTM) \cite{li2019machine}, and Transformer networks \cite{hu2021detection}, can be applied to process the social media data. 
However, such task-specific models have several limits. 
First, task-specific models apply supervised learning to train the models, requiring high-quality labeled data to achieve good performance. Unfortunately, annotating a large dataset on social media can be challenging and time-consuming. 
Second, as platforms improve their detection capabilities, drug dealers constantly adapt their text-based trafficking techniques to avoid detection. This poses a substantial challenge for supervised learning models, as they may struggle to decipher disguised language and euphemisms, potentially resulting in the oversight of crucial information related to drug trafficking activities. 

Meanwhile, recent years have observed the emergence of powerful large language models (LLMs) such as GPT \cite{brown2020language} and LLaMA \cite{touvron2023llama}. These models, trained on huge datasets, have demonstrated surprising capabilities in various natural language processing (NLP) tasks,  such as natural language understanding, generating coherent and contextually relevant responses, and solving complex tasks through text generation. LLMs have shown promise in helping with various real-world tasks, including complex math problems \cite{drori2022neural}, clinical decision support \cite{gpt:jamia2023}, public health \cite{biswas2023role}, open education \cite{firat2023chat}, and global warming \cite{biswas2023potential}. Unlike the supervised learning paradigm, which requires large labeled data to train a task-specific model, LLMs can perform NLP tasks with just a few or no examples, achieving results similar to those of state-of-the-art supervised models \cite{wei2022emergent}. Therefore, LLMs provide an alternative solution to detect illicit drug trafficking on social media.

Inspired by the advance in large language models (LLMs), in this paper, we propose to apply LLMs to detect illicit drug trafficking activities from text data on social media (e.g.,  textual information of posts on Instagram).  Although images often go with text data on social media, previous studies \cite{hu2021detection,hu2021identifying} show that text data dominate the detection of illicit drug trafficking activities. Unlike supervised learning that trains a model with input-label pairs, the LLM-based approach performs a downstream task by reformulating the task with an appropriate textual \emph{prompt}, which bridges the task and the LLMs to enable in-context learning in an autoregressive manner \cite{brown2020language}. To apply LLMs for prediction tasks, we must modify the original text into a textual prompt \cite{Liu:Promptsurvey2023}. Consequently, the problem of LLM-based drug trafficking detection boils down to the design of appropriate prompts.


To this end, we propose an analytical framework to compose \emph{knowledge-informed prompts}, which serve as the interface that humans can interact with and use LLMs for the detection of drug trafficking activities. 
The prompt is the combination of knowledge, question, and original text to detect. 
To construct meaningful prompt knowledge, our proposed framework  integrates  \emph{prior domain knowledge} regarding drug trafficking behaviors, terminologies, and strategies used by drug dealers, and \emph{acquired knowledge} extracted from an LLM, ChatGPT in particular, with a few examples. 
We further design an optimization method to optimize the prompts. The prompt optimization use an iterative strategy \cite{shin2020autoprompt}  to elicit more accurate factual knowledge about drug trafficking than manually created prompts on the benchmark. 
Meanwhile, Monte Carlo dropout \cite{goel2021robustness} is applied to effectively incorporate prior knowledge specific to drug trafficking behaviors than supervised relation extraction models. 

Finally, we assess our proposed framework through
extensive experiments on an illicit drug  trafficking dateset we collected from Instagram. 
Our framework demonstrates superior performance compared to baseline models.
Furthermore,  through conducting thorough evaluations, including ablation experiments and assessments with varying input shots, we demonstrate the importance of prompt design, the value of domain knowledge integration, and the optimal threshold of input shots to maximize performance.

In summary,  our research makes the following contributions:

\begin{itemize}
\item We conduct the first systematic study on leveraging large language models (LLMs), such as ChatGPT,  to detect illicit drug trafficking activities on social media. 
\item We propose an analytical framework to compose knowledge-informed prompts, which serve as the interface that humans can interact with and use LLMs for the detection of drug trafficking activities.
\item We design a prompt optimization method to further enhance  the effectiveness of the prompts. 
\item We demonstrate the effectiveness of the proposed framework through extensive experiments on an illicit drug  trafficking dateset we collected from Instagram. 

\end{itemize}

Our research contributes to the broader objective of improving online security and public safety. Through the integrated prompts of prior knowledge and ChatGPT knowledge, our approach provides an effective tool to detect drug trafficking activity on social media.  Our research offers a practical and applicable tool for law enforcement agencies, social media managers, and other stakeholders concerned with online security and public safety.

The remainder of this paper is organized as follows: Section 2 provides an overview of related work in drug trafficking detection and large language models, Section 3 describes the proposed method in detail, Section 4 presents experimental results and analysis, and Section 5 concludes the article while outlining potential future avenues of research.

\section{Related Work}\label{sec:2}

\subsection{Drug trafficking detection on social media}

As social media platforms have emerged critical channels for drug marketing and illegal sale,  how to effectively and efficiently detect  illicit drug trafficking activities becomes important in addressing this issue. 
There are a few studies that build task-specific machine learning models to detect drug trafficking activities on social media, including detection of drug dealers \cite{hu2021identifying,li2019machine}, drug trafficking events \cite{yang2017tracking, hu2021detection,ZhaoTOCBB22}, and drug-related hashtags \cite{hu2023fine}. 
For example, Li  et al. \cite{li2019machine} applied long short-term memory networks (LSTM) to process textual data of Instagram posts to detect and characterize  illicit drug dealers on Instagram. Similarly, Zhao et al. \cite{ZhaoTOCBB22} combined both SVM and TextCNN to detect illicit drug ads. 
Since social media posts contain both textual and image data, there are some research on apply multimodal approach to  drug trafficking detection \cite{hu2021identifying,hu2021detection,yang2017tracking}. 
Qian et al.  \cite{qian2021distilling} employed heterogeneous graph to capture multi-modal content and relational structured information from social media to detect illicit drug traffickers. More recently, Hu et al. \cite{hu2023fine} proposed a framework that combined Bidirectional Encoder Representations from Transformers (BERT) with Graph Convolutional Network (GCN) to classify drug-related hashtags on Instagram. 

Existing studies that built task-specific models for drug trafficking detection have several limits. 
First,  they need large amount of high-quality labeled data to train the models, however, data annotation is time-consuming and resource-intensive. 
Second,  as drug dealers constantly update their deceptive language and euphemisms to avoid detection, a well-trained model might fail to perform the detection task. 
As large language models  advance, the challenges associated with drug trafficking detection in social media are expected to be effectively addressed.

\subsection{Large language models}
Large language model (LLMs) are  sophisticated artificial intelligence models trained on vast amounts of text data and demonstrate advanced language processing capabilities \cite{zhao2023LLMsurvey}. 
Several notable large language models have been developed, including OpenAI's GPT (Generative Pre-trained Transformer) series and Google's BERT (Bidirectional Encoder Representations from Transformers). 
These large language models are typically pre-trained on massive datasets, often comprising billions of sentences from various sources such as books, websites, and articles. During pre-training, the models learn to predict missing words in sentences, thereby gaining a deep understanding of grammar, context, and semantic relationships. One prominent example of a large language model is ChatGPT \cite{brown2020language} developed by OpenAI. 

Once pre-training is completed, LLMs can  can perform  various natural language processing (NLP) tasks with just a few or no examples, achieving results similar to those of state-of-the-art supervised models \cite{wei2022emergent}. LLMs also have shown promise in helping with various real-world tasks, including complex math problems \cite{drori2022neural}, clinical decision support \cite{gpt:jamia2023}, public health \cite{biswas2023role}.  


\subsection{Prompt engineering based on language models}
Prompts are the interface for humans to interact with and use LLMs.  LLM-based approach performs a downstream task by reformulating the task with an appropriate textual prompt, which bridges the task and the LLMs to enable in-context learning in an autoregressive manner \cite{brown2020language}.
Prompt engineering \cite{Liu:Promptsurvey2023} has  received increasing attention in recent years. Researchers have explored different methods to design prompts that can improve the performance of language models on specific tasks. For example, Shin et al. \cite{shin2020autoprompt} proposed a method for designing prompts to improve the performance of language models on various natural language processing tasks. Their approach, AutoPrompt, leverages a small set of labeled examples to automatically generate prompts that guide the language model toward the desired task. They achieved significant performance gains across various benchmarks by fine-tuning the model with task-specific prompts. A proposed method facilitates a chain-of-thought prompting approach, enabling expansive language models to tackle intricate reasoning tasks by generating a sequence of intermediate steps \cite{wei2022chain}. This methodology sheds light on the emergent property of model scale while expanding the repertoire of reasoning tasks language models can proficiently undertake.  An interactive system called PromptChainer has been developed to facilitate the visual programming of LLM chains, empowering individuals without AI expertise to prototype AI-infused applications \cite{wu2022promptchainer}. However, these methods fail to take advantage of domain knowledge to enhance the performance of the large-language model.


\section{Methodology}\label{sec:3}

In this section, we first formulate the problem of drug trafficking detection on social media, and then introduce the knowledge-prompted large-language model approach to the problem.

\subsection{Problem Formulation}\label{sec:3.1}

\begin{figure}
 \centering
 \includegraphics[width=0.8\linewidth]{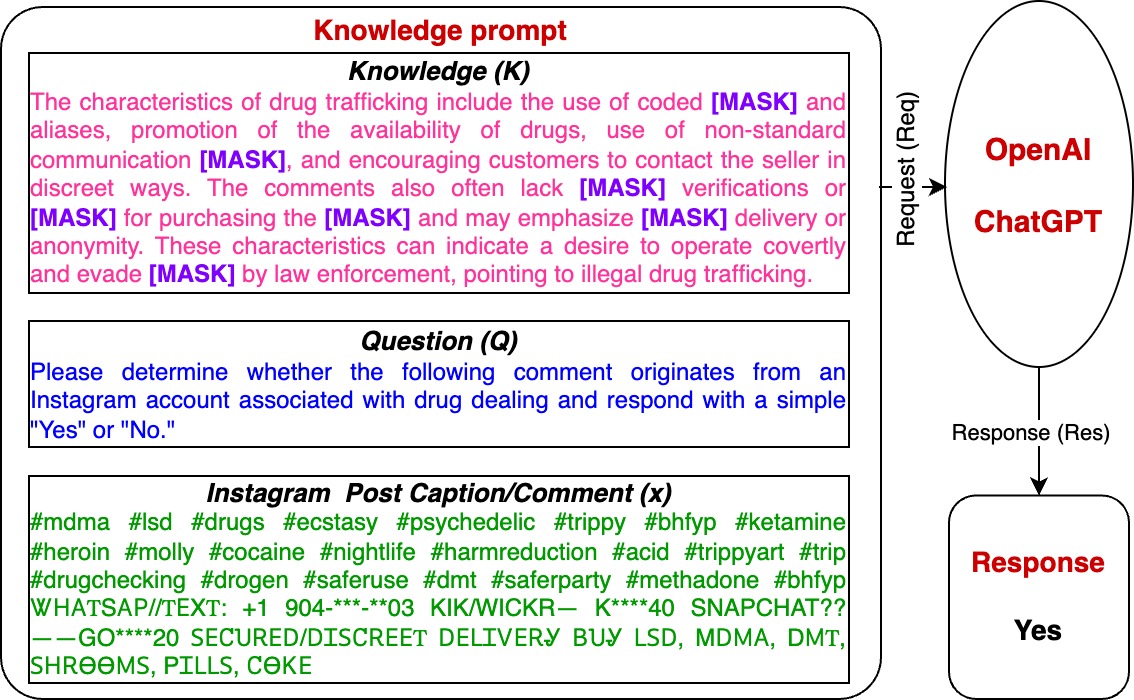}
 \caption{Illustration of the proposed \emph{knowledge-informed prompt}, which is the interface for humans to interact with and use ChatGPT for drug trafficking detection. The prompt $x': [K, Q, x]$ is a tuple of   knowledge $K$, question $Q$, and original  text $x$. 
 The pink font represents knowledge $K$ with  masked words in purple, the blue font represents question $Q$, green font represents original  text $x$ (e.g., Instagram post captions/comments).}
  \label{fig:0}
\vspace{-10pt}
\end{figure}

Our goal is to design an effective system to detect drug trafficking activities on social media platforms accurately. Specifically, our objective is to classify whether a social media post (e.g., a post on Instagram or a tweet on Twitter) contains information related to the marketing and sales of illicit drugs so that the classification can be applied to monitor drug trafficking activities on social media platforms. In this paper, we use only the textual data from the posts. However, images often go with text on social media, but textual data have a dominating role in detecting illicit drug trafficking activities \cite{hu2021detection,hu2021identifying}. 
Let $x$ denote the textual data in a post, our goal is to build a predictive model that takes $x$ as input and predicts a drug trafficking label $y\in \{0, 1\}$, where $y=1$ denotes the post is related to drug trafficking activities and $y=0$ otherwise. 
The block highlighted in green in Figure \ref{fig:0} shows an example of the textual data of an Instagram post annotated as a drug trafficking post. 
In particular, we propose to leverage large language models (LLMs), such as ChatGPT, so that we can 
\begin{enumerate}[(1)]
    \item To leverage the rich knowledge of LLMs acquired through training with large datasets, and
    \item To enable the detection with a few (i.e., few-shot learning where the LLM model is given a few demonstrations of the task) or even no (i.e., zero-shot learning) examples. 
\end{enumerate}

Different supervised learning, which trains a model with large size of input-label pairs $\{(x_i, y_i)\}_{i \in [N]}$, we need to modify the original  text $x$  into a textual \emph{prompt} $x'$ to perform the drug trafficking task. However, how to design appropriate prompts is not a nontrivial task, as the prompts have an important impact on the effectiveness of LLMs. As a result, the problem of LLM-based drug trafficking detection boils down to the design of appropriate prompts, which is the focus of this paper.

\subsection{Overview of Proposed Framework}

As shown in Figure \ref{fig:0}, our proposed \emph{knowledge-informed prompt} $x': [K, Q, x]$ is composed of   knowledge $K$, question $Q$, and original text $x$. 
Then the knowledge-informed prompt $x'$ is the interface that humans can interact with and use ChatGPT for the detection of drug trafficking.  ChatGPT utilizes this prompt to generate responses that help detect drug trafficking activities. The response component represents the system output, providing valuable insights and detection results.

\begin{figure}[tbh]
 \centering
 \includegraphics[width=0.65\linewidth]{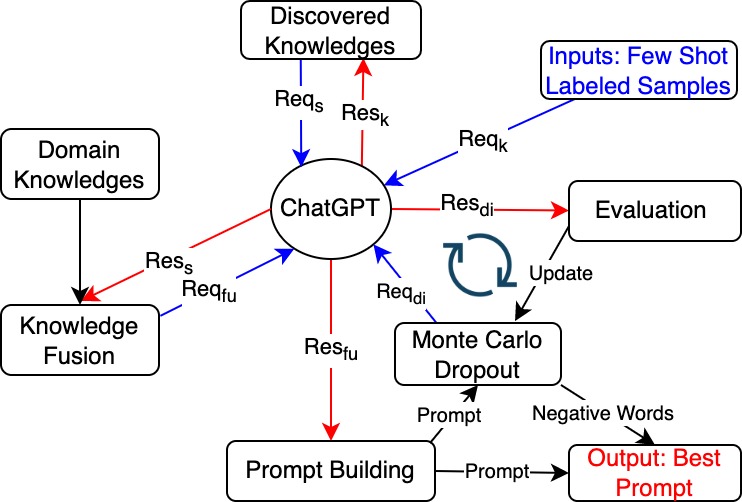}
 \caption{Framework of our proposed knowledge prompted ChatGPT. Blue arrows represent requests to ChatGPT, while red arrows signify the corresponding responses from ChatGPT. Inputs to the framework are indicated by text in blue, whereas outputs are denoted by text in red.}
  \label{fig:1}
\vspace{-10pt}
\end{figure}

\begin{table}
\centering
\small
\caption{List of symbols used to describe the proposed method.}
\label{tab:tab_1}
\begin{tabular}{|l|l|}
\hline
\multicolumn{1}{|c|}{\textbf{Notation}} & \multicolumn{1}{c|}{\textbf{Meaning}}                                                                                                                 \\ \hline
$Req_{k}$                               & Request knowledge discovery based on a few-shot Instagram comments.                                                                                   \\ \hline
$Res_{k}$                               & Response extracted knowledge from a few-shot Instagram comments.                                                                                      \\ \hline
$Req_{s}$                               & Request the combination and summarization of the extracted knowledge.                                                                                 \\ \hline
$Res_{s}$                               & Response that combines and summarizes the extracted knowledge.                                                                                        \\ \hline
$Req_{fu}$                              & Request the fusion of extracted knowledge and domain knowledge.                                                                                       \\ \hline
$Res_{fu}$                              & Response the fusion of extracted knowledge and domain knowledge.                                                                                      \\ \hline
$Req_{di}$                              & \begin{tabular}[c]{@{}l@{}}Request the detection of Instagram comments using the prompt after\\ randomly dropping out the ith iteration.\end{tabular} \\ \hline
$Res_{di}$                              & Response for the prediction of each input Instagram comment data.                                                                                     \\ \hline
\end{tabular}
\end{table}

To design  effective \emph{knowledge-informed prompts} so as to leverage the capabilities of  large language models (LLMs) for the detection of drug trafficking activities on social media, we propose an advanced analytical framework that integrates  \emph{prior domain knowledge} and \emph{acquired knowledge}  extracted from an LLM, ChatGPT in particular.
As shown in Figure \ref{fig:1}, the framework builds upon ChatGPT as its core, combining the power of the large language model with prior knowledge and acquired knowledge extracted from ChatGPT. Different notations in the framework, as shown in Table \ref{tab:tab_1}, represent distinct meanings and functions in terms of knowledge request and response. This framework encompasses a series of steps to design effective prompts to improve the model's effectiveness in detecting drug trafficking activities.

\begin{enumerate}
\item\textbf{Knowledge extraction from ChatGPT.} In this step, we leverage the capabilities of ChatGPT to extract knowledge with a few shots of labeled data $\{(x_i, y_i)\}_{i \in [N]}$, where $N$ is a small number. We input relevant text passages or queries into ChatGPT and utilize its language comprehension and generation capabilities to extract key facts, insights, and relationships related to drug trafficking activities. Various sources such as news articles, online forums, and social networks are used to extract valuable information.
\item \textbf{Knowledge fusion.} The knowledge extracted from ChatGPT is then integrated with domain-specific knowledge to improve the drug trafficking detection capabilities of the framework. Domain-specific knowledge includes information from experts, research articles, and curated databases, providing a deep understanding of drug trafficking patterns, terminology, smuggling techniques, and key entities involved. The integration process aligns and reconciles the extracted knowledge with the domain knowledge, capturing nuanced insights from both sources.
\item \textbf{Prompt design based on integrated knowledge.} To effectively prompt the ChatGPT model, we design prompts that leverage integrated knowledge and exploit areas of potential confusion. Confusion knowledge refers to specific aspects or concepts that ChatGPT might struggle to understand or disambiguate accurately. By addressing and clarifying these confusion points in the prompts, we guide the model's attention and enhance its understanding of drug trafficking-related text. The prompts may involve specific questions, context framing, or the inclusion of key keywords related to drug trafficking activities.
\item \textbf{Prompt optimization.} Prompt optimization aims to fine-tune the designed prompts to maximize the performance of the ChatGPT model on drug trafficking detection tasks. Monte Carlo drop is employed to iteratively improve the prompts' effectiveness. By analyzing the model's responses to different prompts and measuring their impact on performance, we iteratively refine the prompts to align the model's behavior more closely with the desired drug trafficking detection outcomes.
\end{enumerate}

The pipeline of knowledge extraction, knowledge fusion, prompt design based on confusion knowledge, and prompt optimization is crucial in enhancing the model's understanding and performance in identifying drug trafficking activities. We elaborate the details for each part in subsequent sections.

\subsection{Incorporating Prior Domain Knowledge}
\label{sec:3.3}

Domain-specific knowledge holds an immense importance in improving the detection of drug trafficking activities on social media platforms. Domain experts possess specialized knowledge and insights into drug trafficking behaviors, terminologies, and strategies used by drug dealers. By integrating this knowledge into the detection framework, we can improve the accuracy and effectiveness of identifying drug trafficking.
To leverage domain knowledge, we focus on three key aspects: hashtags, contact information, and special symbols commonly used by drug dealers to avoid detection. Table \ref{tab:2} provides examples and meanings for each aspect.

\begin{table}
\centering
\small
\caption{Types of domain knowledge for   drug trafficking detection.}
\label{tab:2}
\begin{tabular}{|l|l|l|}
\hline
\multicolumn{1}{|c|}{Name}                                    & \multicolumn{1}{c|}{Meaning}                                                                                         & \multicolumn{1}{c|}{Example}                                                                                                                            \\ \hline
Hashtag                                                       & Drug sale-related hashtags                                                                                           & \#MDMA \#Cocaine \#LSD                                                                                                                                  \\ \hline
\begin{tabular}[c]{@{}l@{}}Contact\\ information\end{tabular} & \begin{tabular}[c]{@{}l@{}}Telephone number, email\\ address, and other private\\ social media accounts\end{tabular} & \begin{tabular}[c]{@{}l@{}}Txt/WhatsApp.+1,7**.***.9414 \\ Wichr/snapchat james*****52 \\ kik james*****52\end{tabular}                       \\ \hline
\begin{tabular}[c]{@{}l@{}}Special\\ symbol\end{tabular}      & \begin{tabular}[c]{@{}l@{}}Using Punctuation, \\ special characters, and\\ emojis to evade detection\end{tabular}    & \begin{tabular}[c]{@{}l@{}}M.D.M.A, C.O.C.A.I.N.E, L.s.d \\ M.o.l.l.y, SHR$\Theta$$\Theta$MS\\C$\Theta$KE\end{tabular} \\ \hline
\end{tabular}
\end{table}

\begin{itemize}
\item \textbf{Hashtag:} Drug sale-related hashtags are frequently used on social media platforms to facilitate communication and advertising among drug dealers and potential buyers. By identifying and analyzing these hashtags, we can gain valuable information about drug trafficking activities. For example, hashtags like "\#MDMA" or "\#Cocaine" often indicate the sale or discussion of specific drugs.
\item \textbf{Contact Information:} 
Drug dealers often share contact information, such as phone numbers or alternative social media accounts, to establish communication with potential buyers. By monitoring and analyzing this contact information, we can detect and track drug trafficking activities. Examples of contact information include phone numbers with specific area codes or messaging app usernames like "Telegram" or "Wickr."
\item \textbf{Special Symbols:} 
To evade detection, drug dealers frequently use various techniques, including the use of punctuation marks, special characters, or emojis, to mask content related to drug trafficking. When we recognize and interpret these special symbols, we can reveal the underlying drug-related messages. Examples include variations in drug names using special characters such as "M.D.M.A" or "C.O.C.A.I.N.E," or substituting letters with similar-looking symbols like "SHR$\Theta$$\Theta$MS."
\end{itemize}

Incorporating domain-specific knowledge related to hashtags, contact information, and special symbols provides valuable contextual insights that help uncover drug-trafficking discussions and improve the precision and recall of the detection system. This domain knowledge will be fused with acquired knowledge extracted from ChatGPT to design prompts.

\subsection{Knowledge Fusion and Prompt Engineering}

\begin{figure}[h]
 \centering
 \includegraphics[width=0.99\linewidth]{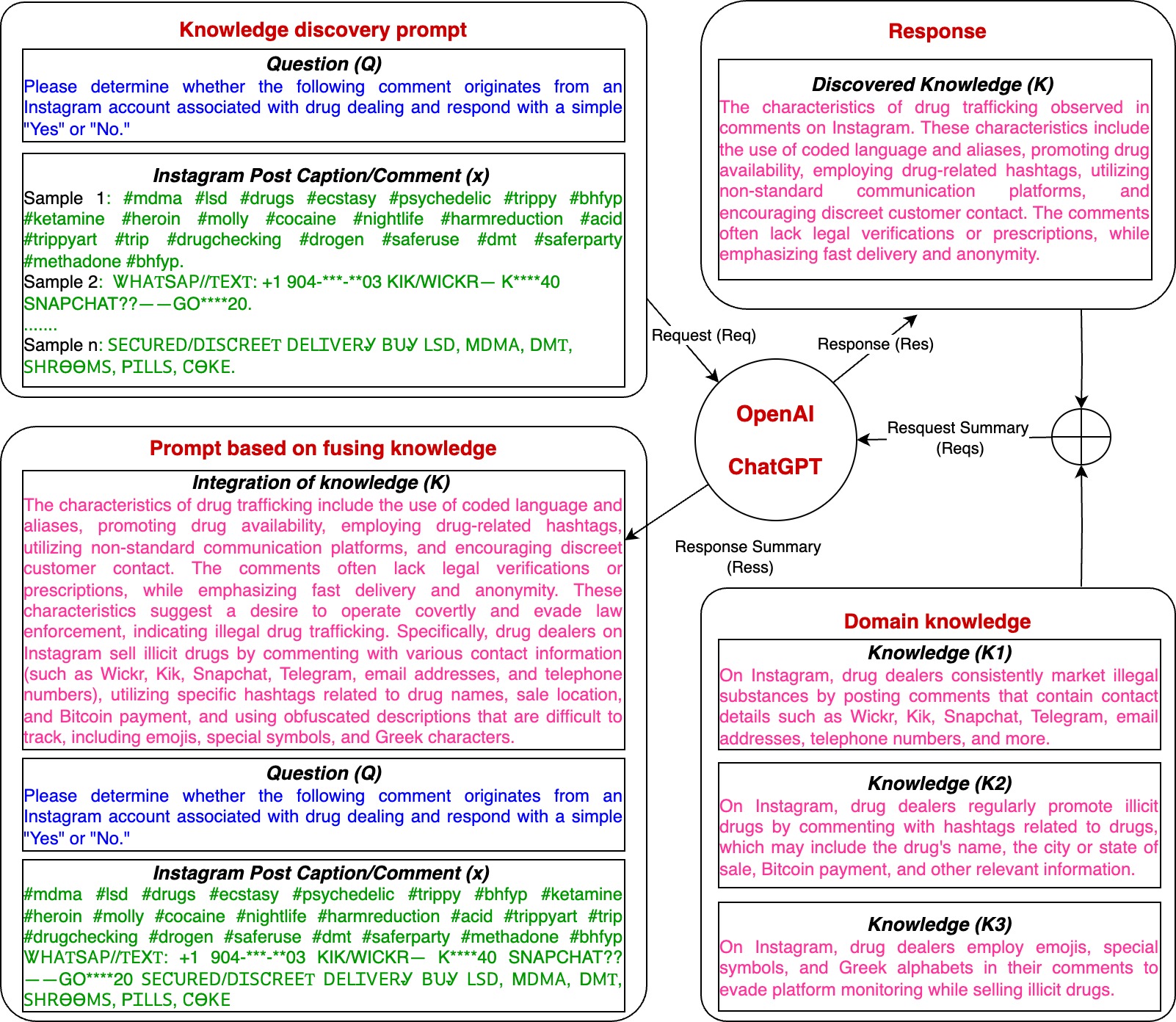}
 \caption{Illustration of the workflow to compose prompt $x': [K, Q, x]$, which is a tuple of   knowledge $K$, question $Q$, and original  text $x$. The pink font represents knowledge $K$, blue font represents question $Q$, green font represents original  text $x$ (e.g., Instagram post captions/comments). }
\label{fig:3}
\end{figure}

Note that given a text data $x$, our proposed \emph{knowledge-informed prompt} $x': [K, Q, x]$ is the combination of  knowledge $K$, question $Q$, and original text $x$. Figure \ref{fig:3} shows the workflow to compose the prompt.

\subsubsection{Prompt -- Knowledge}
In particular, we compose the knowledge part $K$ with \emph{prior domain knowledge}  and \emph{acquired knowledge}  extracted from ChatGPT with a few shots of labeled data.  


{\noindent \textbf{Knowledge Discovery with ChatGPT.}} 
Beyond the \emph{prior domain knowledge} as discussed in  Section \ref{sec:3.3}, we also extract relevant knowledge with  ChatGPT since ChatGPT is trained on very large datasets. Specifically, we query ChatGPT with a few shots of labeled data $\{(x_i, y_i)\}_{i \in [N]}$, where $N$ is a small number, to guide ChatGPT  to generate informative and relevant responses. 
During fine-tuning, the model learns to identify patterns, relationships, and insights within the drug trafficking domain. This allows the model to acquire implicit knowledge from the dataset and generate informed responses when asked for drug-trafficking-related queries.

 {\noindent \textbf{Knowledge Fusion as Prompt.}} Knowledge fusion is a critical component of the prompt design process. It involves integrating relevant information from domain knowledge and knowledge discovered by ChatGPT into the prompts provided to the model. By fusing knowledge into the prompts, we provide the ChatGPT model with additional context and cues related to drug trafficking. This integration helps the model to make more accurate predictions and generate responses that align with the nuances of drug trafficking detection tasks. For example, a prompt could include a snippet of domain knowledge about common drug trafficking routes or the characteristics of illicit drug concealment methods. This prompts the model to focus its attention on these aspects and generate more insightful responses.

\subsubsection{Prompt -- Question}
To further enhance the design of knowledge prompts, we identify specific types of questions that generate informative responses related to the detection of drug trafficking. These questions are designed to target key aspects of drug trafficking activities, methods, or indicators.
For instance, questions like "What are the signs of drug trafficking in online communications?" or "How can drug smuggling be detected at border checkpoints?" guide the model to provide relevant information in its responses.
By incorporating carefully crafted questions into the prompts, we guide the ChatGPT model to generate more focused and informative outputs for drug trafficking detection tasks.


\subsection{Prompt Optimization with Monte Carlo Dropout}\label{sec:3.4}

In this section, we present the methodology for prompt optimization, leveraging the Monte Carlo dropout technique to enhance the performance of the knowledge-prompt design framework for drug trafficking detection.

\begin{algorithm}[tb]
    \caption{Word Importance Score Calculation with Monte Carlo Dropout for Prompt Optimization.}
    \label{alg:3}
    \small
    \begin{algorithmic}[1]
        \Procedure{Calculate Importance Scores}{$\text{paragraph}, \text{num\_iterations}, \text{dropout\_prob}$}
            \State Initialize an empty dictionary $\text{word\_scores}$
            \State Split the $\text{paragraph}$ into a list of words
            \For{$\text{iter}$ \textbf{in} $1$ \textbf{to} $\text{num\_iterations}$}
                \State Create a copy of the list of words called $\text{masked\_words}$
                \State Initialize an empty list called $\text{dropout\_list}$
                \For{$\text{word\_index}$ \textbf{in} $0$ \textbf{to} $\text{length}(\text{masked\_words}) - 1$}
                    \If{$\text{random.random()} < \text{dropout\_prob}$}
                        \State Append $\text{word\_index}$ to $\text{dropout\_list}$
                        \State Set $\text{masked\_words[word\_index]}$ as "MASK"
                    \EndIf
                \EndFor
                \State Create a $\text{prompt\_knowledge}$ by joining $\text{masked\_words}$ into a string
                \State Perform Monte Carlo on $\text{prompt\_knowledge}$ to get performance metrics
                \State Compute the change in $\text{score\_change}$
                \For{$\text{word\_index}$ \textbf{in} $\text{dropout\_list}$}
                    \If{$\text{word\_index}$ not in $\text{word\_scores}$}
                        \State Set $\text{word\_scores[word\_index]}$ as $0.0$
                    \EndIf
                    \State Increment $\text{word\_scores[word\_index]}$ by $\text{score\_change}$
                \EndFor
            \EndFor
            \State Sort $\text{word\_scores}$ by value in descending order, resulting in $\text{sorted\_scores}$
            \State \textbf{return} $\text{sorted\_scores}$
        \EndProcedure
    \end{algorithmic}
\end{algorithm}

The prompt optimization process aims to fine-tune the designed prompts to maximize the performance of the ChatGPT model on drug trafficking detection tasks. We can identify keywords that significantly influence the model's decisions by systematically analyzing the impact of individual words in the prompt. The prompt optimization is shown in Algorithm \ref{alg:3}.

The algorithm, "Prompt Optimization with Monte Carlo Dropout," takes the designed prompt, the number of Monte Carlo dropout iterations, and the dropout probability as input. The algorithm iteratively performs the following steps:
\begin{itemize}

\item Create a copy of the designed prompt, optimized\_prompt, to preserve the original prompt (Step 1).
\item Split optimized\_prompt into a list of words.
\item For each word in the prompt, perform the Monte Carlo dropout by replacing the word with a "MASK" token (Steps 4-6).
\item Perform Monte Carlo simulation on the modified prompt to obtain performance metrics.
\item Calculate the change in F1 score, score\_change, based on the original and simulated F1 scores.
\item Update the importance score of each word in the prompt by incrementing it with score\_change.
\item Repeat Steps 2-6 for the desired number of iterations.
\end{itemize}
By optimizing the prompt using Monte Carlo dropout, we can identify and refine the words that significantly impact the model's performance in drug trafficking detection tasks.

The resulting optimized prompt provides valuable insight into the specific words and their contributions to the model's decision-making process. This enhanced understanding enables us to fine-tune the prompt and align the model's behavior more closely with the desired drug trafficking detection outcomes.

\section{Experiments}\label{sec:4}

\subsection{Experimental Settings}\label{sec:4.1}

In this subsection, we describe the experimental setup used to evaluate the performance of the knowledge-prompted ChatGPT framework for drug trafficking detection. The experimental setup encompasses the dataset, model configuration, evaluation metrics, and baselines.

\begin{itemize}

\item \textbf{Dataset.} To evaluate the framework's effectiveness, we carefully selected a comprehensive dataset called IDDIG that contains text samples related to drug trafficking on Instagram. Similar samples were removed from the IDDIG data, resulting in a final set of 886 samples for the evaluation experiments. There are 486 positive samples (i.e., drug trafficking) and 400 negative samples (i.e., non-drug trafficking). We propose a framework for identifying positive samples. The dataset consists of labeled data for validation and testing, facilitating the computation of performance metrics. It is thoughtfully curated to encompass various drug trafficking scenarios, regions, and languages, ensuring the framework's generalizability. Specifically, a random set of 40 samples was selected as input for the design of knowledge prompts in each experiment. In contrast, the remaining samples were set aside as a test set to assess the performance of the proposed method.

\item \textbf{Model Configuration.} Our framework is based on the ChatGPT API, specifically using the GPT Turbo 3.5 model. This model undergoes pre-training on a significant corpus of text data to establish a robust language understanding. To accurately detect drug trafficking, we optimize the ChatGPT request prompt by incorporating the knowledge prompt design and optimization techniques mentioned earlier. The model configuration involves several parameters, such as the number of iterations for word dropout, the number of input samples, and the number of words to be dropped out in each iteration. Parameter tuning can be performed to enhance the model's performance.

\item \textbf{Evaluation Metrics.} To evaluate the performance of the knowledge-prompted ChatGPT framework, various metrics have been used. The chosen metrics encompass precision, recall, F1 score, and accuracy. Precision evaluates the accuracy of identifying drug trafficking instances, while recall measures the effectiveness of correctly identifying actual drug trafficking instances. The F1 score combines precision and recall, offering a balanced measure of the model's performance.

\item \textbf{Baselines.} To compare the performance of the knowledge-prompted ChatGPT framework, it is important to establish appropriate baselines. Baselines include BERT \cite{devlin2018bert}, XLNet \cite{yang2019xlnet}, ALBERT \cite{lan2019albert}, DistilBERT \cite{sanh2019distilbert}, and RoBERTa \cite{liu2019roberta} for the detection of drug trafficking. These baselines provide reference points to evaluate the effectiveness and superiority of the proposed framework. Baseline models are trained and evaluated on the same dataset as the knowledge-prompted ChatGPT framework, using the same evaluation metrics to ensure a fair comparison.

\end{itemize}


\subsection{Evaluation of knowledge-prompted ChatGPT for drug trafficking detection}\label{sec:4.2}

\begin{table}[tb]
\centering
\caption{Performance Metrics of Different Models}
\label{tab:4}
\begin{tabular}{|l|c|c|c|c|}
\hline
\textbf{Model} & \textbf{Accuracy} & \textbf{Precision} & \textbf{Recall} & \textbf{F1 Score} \\ 
\hline
BERT \cite{devlin2018bert}            & 76.23\%           & 75.39\%            & 85.91\%         & 79.78\%           \\ 
XLNet \cite{yang2019xlnet}            & 76.13\%           & 75.44\%            & 86.76\%         &  79.92\%           \\ 
ALBERT \cite{lan2019albert}            & 77.62\%           & 78.06\%            & 83.14\%         & 80.13\%           \\ 
DistilBERT \cite{sanh2019distilbert}      & 80.30\%           & 80.73\%            &  86.03\%         & 82.68\%           \\ 
RoBERTa \cite{liu2019roberta}        & 82.78\%           & 83.29\%            & 88.29\%         & 85.02\%           \\ 
\hline
Ours            & \textbf{94.58\%}           & \textbf{96.60\%}            &  \textbf{93.42\%}         & \textbf{94.98\%}           \\ 

\hline
\end{tabular}
\end{table}

This section presents the evaluation results of our knowledge-prompted ChatGPT framework for drug trafficking detection. We compare the performance of our framework with several baseline models, including BERT, XLNet, ALBERT, DistilBERT, and RoBERTa. The evaluation metrics include accuracy, precision, recall, and F1 score (See Table \ref{tab:4}).

We noted a significant enhancement across all evaluation metrics after comparing the outcomes with the baseline models. Our framework demonstrated superior performance in drug trafficking detection tasks by achieving higher accuracy, precision, recall, and F1 scores than all other models. Notably, the proposed framework exhibited exceptional improvements, with nearly a 12\% increase in drug trafficking detection accuracy and an approximately 10\% improvement in F1 score, surpassing other baseline language models.

The significant performance gains can be attributed to incorporating domain-specific knowledge, effective prompt design, and knowledge optimization techniques employed in our framework. The integration of relevant knowledge and the guidance provided through carefully designed prompts contribute to the framework's ability to detect drug trafficking activities accurately.

\subsection{Assessing the performance of knowledge prompt ChatGPT with varying input shots}\label{sec:4.3}
To evaluate the impact of varying input shots on the performance of the knowledge-driven ChatGPT framework for drug trafficking detection, we conducted experiments using different numbers of input shots. The evaluation metrics include accuracy, precision, recall, and F1 score.
\begin{table}[tb]
\centering
\caption{Performance Metrics of Different Models}
\label{tab:performance}
\begin{tabular}{|l|l|l|l|l|}
\hline
\textbf{Shot number} & \textbf{Accuracy} & \textbf{Precision} & \textbf{Recall} & \textbf{F1 Score} \\ 
\hline
5 shots     & 86.23\%           & 90.63\%            & 83.54\%         & 86.94\%           \\ 
10 shots     & 88.71\%           & 91.77\%            & \textbf{87.24\%}         & 89.45\%           \\ 
20 shots      & 87.81\%           & 90.91\%            & 86.42\%         & 88.61\%           \\ 
30 shots          & 87.13\%           & 95.15\%            & 80.66\%         & 87.31\%           \\ 
\hline
40 shots   & \textbf{89.84\%}           & \textbf{95.41\%}            &  85.60\%         & \textbf{90.24\%}         \\   
\hline

\end{tabular}
\end{table}

The framework's performance consistently improved with an increasing number of input shots. Using 40 shots yielded the best results, achieving an accuracy of 89.84\% and an F1 score of 90.24\%, while using five shots resulted in an accuracy of 86.23\% and an F1 score of 86.94\%. These results indicate that the knowledge-driven ChatGPT framework benefits from increasing input shots up to a certain point, leading to improved performance in drug trafficking detection tasks. However, after reaching the optimal point (in this case, around 40 shots), further increases in input shots may not necessarily yield significant performance gains.

\subsection{Evaluating Model Components through Ablation Experiments}\label{sec:4.4}

To assess the individual contributions of different model components in the knowledge-driven ChatGPT framework for drug trafficking detection, we conducted ablation experiments. We evaluated the performance when removing or modifying specific prompt sources and knowledge components. The evaluation metrics include accuracy, precision, recall, and F1 score. The evaluation results are summarized in Table \ref{tab:6}.

\begin{table}[tb]
\centering
\small
\caption{Performance Metrics of Different Models}
\label{tab:6}
\begin{tabular}{|l|l|l|l|l|l|}
\hline
\textbf{Prompt source} & \textbf{knowledge} & \textbf{Accuracy} & \textbf{Precision} & \textbf{Recall} & \textbf{F1 Score} \\ 
\hline
No prompt & No knowledge &  87.81\%           & 90.21\%            & 87.24\%         & 88.70\%           \\ 
Domain expert & Hashtag           & 87.81\%           & 92.76\%            & 84.36\%         & 88.36\%           \\ 
Domain expert & Contact information         & 87.36\%           & 97.46\%            & 79.01\%         & 87.27\%           \\ 
Domain expert & Special symbol        & 90.52\%           & 89.11\%            & \textbf{94.24\%}         & 91.60\%           \\ 
Domain expert &  All Knowledge (AK)   & 90.74\%            & 98.10\%   & 84.77\%      & 90.95\%           \\ 
\hline
Ours w/o dropout &  40 shots   & 89.84\%           & 95.41\%            &  85.60\%         & 90.24\%         \\ 

Ours w/o dropout &  40 shots + AK    & 91.87\%           & \textbf{99.52\%}            &  85.60\%         & 92.04\%           \\ 
Ours w/ dropout      & 40 shots + AK     & \textbf{94.58\%}           & 96.60\%            &  93.42\%         & \textbf{94.98\%}           \\ 
\hline
\end{tabular}
\end{table}

Removing the prompt knowledge components, resulting in no prompt and no knowledge, led to an accuracy of 87.81\%, and an F1 score of 88.70\%. When the domain expert provided a drug-related hashtag as the knowledge prompt source, the performance remained consistent with an accuracy of 87.81\%. Still, the precision increased to 92.76\%, albeit with a slightly lower recall of 84.36\% and an F1 score of 88.36\%. Using the domain expert's contact information as the prompt source improved precision significantly to 97.46\%. However, this change resulted in a decreased recall of 79.01\% and a slightly lower accuracy of 87.36\%, with an F1 score of 87.27\%. Replacing the prompt source with a special symbol guided by the domain expert yielded an accuracy of 90.52\%, precision of 89.11\%, recall of 94.24\%, and an F1 score of 91.60\%. These results indicate that the special symbol prompt source effectively captures the model's attention and improves its ability to identify drug trafficking activities. When all knowledge components provided by the domain expert were included, the performance further improved. The framework achieved an accuracy of 90.74\% and an F1 score of 90.95\%. This highlights the significance of incorporating comprehensive domain knowledge to enhance the framework's performance.

Removing prompt knowledge components resulted in an accuracy of 87.81\% and an F1 score of 88.70\%. Utilizing a drug-related hashtag as the knowledge prompt source maintained an accuracy of 87.81\%, but precision increased to 92.76\% with a slightly lower recall of 84.36\% and an F1 score of 88.36\%. Using the domain expert's contact information as the prompt source significantly improved precision to 97.46\%, albeit with decreased recall of 79.01\%, accuracy of 87.36\%, and an F1 score of 87.27\%. Replacing the prompt source with a special symbol guided by the domain expert yielded an accuracy of 90.52\%, precision of 89.11\%, recall of 94.24\%, and an F1 score of 91.60\%. Including all knowledge components provided by the domain expert further enhanced performance, resulting in an accuracy of 90.74\% and an F1 score of 90.95\%. These results demonstrate the effectiveness of different prompt sources and highlight the importance of incorporating comprehensive domain knowledge to enhance the framework's performance in drug trafficking detection.

Next, we examine the effect of Monte Carlo dropout and the number of input shots. Without dropout, our framework with 40 shots achieved an accuracy of 89.84\%, precision of 95.41\%, recall of 85.60\%, and an F1 score of 90.24\%. However, the performance significantly improved when incorporating dropout with the same number of shots. The framework achieved an accuracy of 91.87\%, precision of 99.52\%, recall of 85.60\%, and an F1 score of 92.04\%. Finally, our complete framework with 40 shots and all knowledge components achieved the best performance, with an accuracy of 94.58\%, precision of 96.60\%, recall of 93.42\%, and an impressive F1 score of 94.98\%.

These experiments demonstrate the significance of prompt sources, knowledge components, and the utilization of Monte Carlo dropout in enhancing the accuracy and effectiveness of our framework for drug trafficking detection. The comprehensive integration of domain knowledge and the optimization of model components contribute to the superior performance of our proposed approach.

\subsection{Drug Trafficking Detection Case Studies}
\subsubsection{Leveraging Monte Carlo Dropout for a Knowledge Prompt Example}

\begin{table}[tb]
\centering
\small
\caption{Example Prompts by the proposed method.Green font Represents Positive words, and Red font Represents Negative words.}
\label{tab:7}
\begin{tabular}{|p{8.7cm}|p{1cm}|p{2.1cm}|p{1.8cm}|}
\hline
\textbf{Prompt} & \textbf{TopK} & \textbf{P Words} & \textbf{N Words} \\
\hline
The characteristics of \textcolor{green}{drug} trafficking observed in comments on \textcolor{red}{Instagram}. These characteristics& Top1 & addresses & emojis \\
include the use of coded language and aliases, \textcolor{red}{promoting} drug availability, \textcolor{red}{employing} drug-related & Top2 & delivery & Instagram \\
hashtags, utilizing \textcolor{green}{non-standard} communication platforms, \textcolor{red}{and} encouraging discreet customer & Top3 & enforcement & contact \\
\textcolor{red}{contact}. The comments often lack legal verifications or prescriptions, while emphasizing fast & Top4 & drug & employing  \\
\textcolor{green}{delivery} and \textcolor{red}{anonymity}. These characteristics suggest a desire to \textcolor{green}{operate} covertly and \textcolor{green}{evade} law & Top5 & operate & payment \\
\textcolor{green}{enforcement}, indicating illegal drug \textcolor{green}{trafficking}. Specifically, drug dealers on  \textcolor{green}{Instagram} sell  & Top6 & as & Wickr  \\
illicit drugs by commenting with various contact information (such \textcolor{green}{as} \textcolor{red}{Wickr}, Kik, Snapchat, & Top7 & evade & anonymity \\
email \textcolor{green}{addresses}, and telephone numbers), utilizing specific hashtags related to drug names, sale  & Top8 & Instagram & promoting \\
location, and Bitcoin \textcolor{red}{payment}, and using obfuscated descriptions that are difficult to track,  & Top9 & trafficking & including \\
\textcolor{red}{including} \textcolor{red}{emojis}, special symbols, and Greek characters. & Top10 & on-standard & and \\
\hline
\end{tabular}
\end{table}

In this subsection, we present case studies to demonstrate the effectiveness of our proposed method for drug trafficking detection. The case studies involve the use of positive words and negative words as prompts, highlighting the top 10 words ranked by their calculated importance scores (See Table \ref{tab:7}). 

The prompts are designed to capture the characteristics of drug trafficking discussions observed in comments on Instagram. The identified positive words, represented in green font, indicate significant features associated with drug traffickings, such as drug names, delivery, non-standard communication platforms, and evasion of law enforcement. For example, "address" and "delivery" are considered strong positive words that express key features of drug trafficking. These words play a crucial role in improving the performance of ChatGPT in the detection of drug trafficking. The presence of "address" signifies the use of email addresses as a means of communication, while "delivery" emphasizes fast delivery. These features may have been missing in ChatGPT's understanding of drug trafficking, and their inclusion enhances the model's ability to identify drug trafficking activities accurately. On the contrary, the negative words, represented in red font, signify the words that negatively affect the performance of ChatGPT in detecting drug trafficking activities. These words hinder the model's ability to detect illicit drug-related content on social media platforms accurately. To enhance the performance of drug trafficking detection, negative words will be removed from the analysis. 

The calculated importance scores provide information on the relevance and significance of each word on the detection of drug trafficking. The top-ranked words offer valuable clues to identify and monitor drug trafficking activities on social media platforms. By utilizing these prompts, our method enables a more targeted and effective approach to detecting drug-related content and identifying potential drug dealers.

The case studies highlight the capabilities of our proposed method in capturing key characteristics of drug trafficking discussions, shedding light on the covert nature of such activities, and providing law enforcement agencies and social media managers with valuable insights for intervention and prevention efforts.

\subsubsection{A Comparative Analysis: Proposed Framework vs. Alternative Prompts}

In this subsection, we present a comparative analysis of our proposed method for the detection of drug trafficking, contrasting it with alternative prompts. There are three alternative prompts: a prompt with no specific topic, a prompt focused on domain knowledge, and a prompt based on extracted knowledge.  
\begin{itemize}
    \item \textbf{Prompt a.} No prompt.
    \item \textbf{Prompt b.} Drug dealers on Instagram utilize various methods to sell illicit drugs. They typically comment with contact information (such as wickr, Kik, snapchat, telegram, or telephone number), employ specific hashtags (including names, sale locations, or bitcoin payment), and use obfuscated descriptions with emojis, special symbols, and Greek characters to evade tracking, effectively showcasing the drugs they have for sale.
    \item  \textbf{Prompt c.} The characteristics of drug trafficking include using coded language and aliases, promoting the availability of drugs, using non-standard platforms, and encouraging customers to contact the seller discreetly. The comments also often lack legal verifications or prescriptions for drug purchasing and may emphasize fast delivery or anonymity. These characteristics can indicate a desire to operate covertly and evade detection by law enforcement, pointing to illegal drug trafficking.
    \item  \textbf{Prompt d.} The characteristics of drug trafficking include using coded language and aliases, promoting drug availability, employing drug-related hashtags, utilizing non-standard communication platforms, and encouraging discreet customer contact. The comments often lack legal verifications or prescriptions while emphasizing fast delivery and anonymity. These characteristics suggest a desire to operate covertly and evade law enforcement, indicating illegal drug trafficking. Specifically, drug dealers on Instagram sell illicit drugs by commenting with various contact information (such as Wickr, Kik, Snapchat, Telegram, email addresses, and telephone numbers), utilizing specific hashtags related to drug names, sale location, and Bitcoin payment, and using obfuscated descriptions that are difficult to track, including emojis, special symbols, and Greek characters.
\end{itemize}

\begin{table}[tb]
\caption{Contrasting the Proposed Method with Alternative Prompts: A Comparative Analysis. Prompt a. Represents no prompt, Prompt b. Represents Prompt with All Knowledge, Prompt c. Represents Prompt with Extracted Knowledge, Prompt d. Represents Prompt that fusing Prompt b. and Prompt c. "P" represents drug trafficking label or prediction, while "N" represents non-drug trafficking. Red fonts indicate wrong predictions.}
\small
\label{tab:8}
\begin{tabular}{|l|lllllllllll|}
\hline
ID                 & \multicolumn{11}{c|}{Cases}                                                                                                                                                                                                                                                                                                                                                                                                                                                                                                    \\ \hline
\multirow{2}{*}{1} & \multicolumn{11}{l|}{\begin{tabular}[c]{@{}l@{}}	WH$\alpha$TS$\alpha$P // TEXT +1 760*****43 $\omega$ICKR// KIK //Sn$\alpha$pch$\alpha$t —X****ver \\ SECURED/DISCREET DELIVERY BUY LSD, MDMA, DMT,SHR$\Theta$$\Theta$MS, \\ PILLS, C$\Theta$KE \& more \#lsd\#shrooms \#shroomforsale \#magicshrooms? \#tabs\end{tabular}}                                                                                                                                                                                                                                                             \\ \cline{2-12} 
\cline{2-12} 
                   & \multicolumn{2}{l|}{Label}                                             & \multicolumn{1}{l|}{P}                        & \multicolumn{1}{l|}{Prompt a.}                        & \multicolumn{1}{l|}{P}                        & \multicolumn{1}{l|}{Prompt b.}                        & \multicolumn{1}{l|}{P}                        & \multicolumn{1}{l|}{Prompt c.}                        & \multicolumn{1}{l|}{P}                        & \multicolumn{1}{l|}{Prompt d.}                        & P                       \\ \hline
\multirow{2}{*}{2} & \multicolumn{11}{l|}{\begin{tabular}[c]{@{}l@{}}HOW TO AVOID RELIANCE ON PILLS ? \#medication Synovation Medical \\ Group offers a Functional Restoration Program where our physicians help \\ patients overcome chronic pain through an interdisciplinary approach so\\ they can return to normal work and life. This program helps -Increasing\\ the reliance on one's self-Decreasing pain medications \#painmedication \\ \#healthcare  \#medicalcare \#chronicpain \#healthiswealth \#opioidcrisis "\end{tabular}} \\ \cline{2-12} 
                   & \multicolumn{2}{l|}{Label}                                             & \multicolumn{1}{l|}{N}                        & \multicolumn{1}{l|}{Prompt a.}                        & \multicolumn{1}{l|}{\color{red}{P}}                        & \multicolumn{1}{l|}{Prompt b.}                        & \multicolumn{1}{l|}{N}                        & \multicolumn{1}{l|}{Prompt c.}                        & \multicolumn{1}{l|}{N}                        & \multicolumn{1}{l|}{Prompt d.}                        & N                       \\ \hline
\multirow{2}{*}{3} & \multicolumn{11}{l|}{\begin{tabular}[c]{@{}l@{}}\#psychedelic \#psy \#psytime \#psychedelicrock \#magicmushrooms \#lsdtabs\\ \#shrooms \#trippyvibes \#mdmaofficials  \#microdot \#purpleacid \#blotteracid\end{tabular}}                                                                                                                                                                                                                                                                                                        \\ \cline{2-12} 

                   & \multicolumn{2}{l|}{Label}                                             & \multicolumn{1}{l|}{P}                        & \multicolumn{1}{l|}{Prompt a.}                        & \multicolumn{1}{l|}{P}                        & \multicolumn{1}{l|}{Prompt b.}                        & \multicolumn{1}{l|}{\color{red}{N}}                        & \multicolumn{1}{l|}{Prompt c}                         & \multicolumn{1}{l|}{P}                        & \multicolumn{1}{l|}{Prompt d.}                        & P                       \\ \hline
\multirow{2}{*}{4} & \multicolumn{11}{l|}{\begin{tabular}[c]{@{}l@{}}M.D.M.A n N.E.M.B.U.T.A.L, K.U.S.H, C.O.C.A.I.N.E,A.C.I.D,(L.s.d n C.a.r.t) \\ Txt/WhatsApp ..+1,*** Telegram:*** Wickr: *** M.Md,L.s.d n M.o.l.l.y ,a.d.i.e.s\end{tabular}}                                                                                                                                                                                                                                                                                                   
\\ \cline{2-12} 

                   & \multicolumn{2}{l|}{Label}                                             & \multicolumn{1}{l|}{P}                        & \multicolumn{1}{l|}{Prompt a.}                        & \multicolumn{1}{l|}{P}                        & \multicolumn{1}{l|}{Prompt b.}                        & \multicolumn{1}{l|}{P}                        & \multicolumn{1}{l|}{Prompt c.}                        & \multicolumn{1}{l|}{\color{red}{N}}                        & \multicolumn{1}{l|}{Prompt d.}                        & P                       \\ \hline
\multirow{2}{*}{5} & \multicolumn{11}{l|}{\begin{tabular}[c]{@{}l@{}}Self Medicated ?\#weed  \#marijuana \#cbd \#stoner \#ganja  \#indica  \#sativa \\ \#kush \#maryjane \#dank \#medicalmarijuana  \#bong \#hemp \#stoned  \#bhfyp\end{tabular}}                                                                                                                                                                                                                                                                                                     \\ \cline{2-12} 
                   & \multicolumn{2}{l|}{Label}                                             & \multicolumn{1}{l|}{N}                        & \multicolumn{1}{l|}{Prompt a.}                        & \multicolumn{1}{l|}{N}                        & \multicolumn{1}{l|}{Prompt b.}                        & \multicolumn{1}{l|}{\color{red}{P}}                        & \multicolumn{1}{l|}{Prompt c.}                        & \multicolumn{1}{l|}{N}                        & \multicolumn{1}{l|}{Prompt d.}                        & \color{red}{P}                       \\ \hline
\end{tabular}
\end{table}

Based on the alternative prompts and the proposed prompt (i.e., Prompt d.), we utilize sample data and examine the effectiveness of different prompts in identifying drug trafficking activities, as shown in Table \ref{tab:8}. It illustrates the comparison by providing four different prompts: prompt a, representing no prompt; prompt b, which includes all knowledge; prompt c, comprising extracted knowledge; and prompt d, a fusion of prompt b and prompt c. Each sample in the table is labeled with either a positive (P) or negative (N) indication based on the presence or absence of drug trafficking elements.

In Case 1, ChatGPT-based methods accurately detect drug trafficking despite the presence of contact information (e.g., WhatsApp, Wickr, Kik, Snapchat) and drug-related hashtags suggest potential drug trafficking activity. This showcases the effectiveness of our approach in identifying drug trafficking activities even when perpetrators employ deceptive techniques. However, there are instances where ChatGPT makes mistakes. Case 2, although unrelated to drug trafficking, shares similarities that may lead to erroneous predictions without prompts. This highlights the importance of prompts in guiding the model and minimizing false positives. Interestingly, Case 3 reveals that ChatGPT can also make mistakes even when provided with domain knowledge as a prompt. In some cases, including domain knowledge may introduce noise and negatively impact the model's performance compared to having no prompt at all. This suggests that the prompt design and knowledge extraction must be carefully considered. When Instagram drug trafficking events involve special characters such as punctuation marks, the automatically extracted knowledge may hurt ChatGPT's predictions. These cases emphasize the need for robust methods to handle different types of input data and ensure accurate detection. Prompt d, which includes prompt b (domain knowledge), may sometimes lead to similar mistakes, as observed in Case 5. This highlights the complexity of balancing the inclusion of domain knowledge and avoiding potential pitfalls associated with prompt-based approaches.

These case studies provide valuable insights into the performance of our method, highlighting its ability to distinguish between positive and negative samples based on the prompts used. The results showcase the importance of prompt design and the integration of domain knowledge in enhancing drug trafficking detection mechanisms.

\section{Discussions}\label{sec:5}  
This section discusses the key findings and implications of our research on the knowledge prompted ChatGPT framework for drug trafficking detection. We address the framework's performance, the effectiveness of different model components, and the potential limitations and future directions of the research.

\subsection{Performance of the Framework}
The evaluation results demonstrate the superior performance of our knowledge-prompted ChatGPT framework compared to baseline models. Our framework achieved an accuracy of 94.58\%, precision of 96.60\%, recall of 93.42\%, and an F1 score of 94.98\%. These results highlight the effectiveness of leveraging domain-specific knowledge, well-designed prompts, and prompt optimization techniques in enhancing drug trafficking detection capabilities.

Comparisons with state-of-the-art models such as BERT, XLNet, ALBERT, DistilBERT, and RoBERTa further validate the superiority of our framework. It consistently outperformed these models regarding accuracy, precision, recall, and F1 score, indicating its ability to capture drug trafficking activities accurately and make informed predictions.

\subsection{Effectiveness of Model Components}
The ablation experiments provided valuable insights into the effectiveness of different model components. Removing the prompt source and knowledge components significantly decreased performance, underscoring the importance of incorporating domain knowledge and well-crafted prompts. Our framework's ability to fuse domain knowledge into the prompts, guided by request-related questions and knowledge optimization techniques, played a crucial role in achieving the observed performance gains.

The analysis of varying input shots showed that the framework's performance improved with increasing shots up to an optimal point. Beyond that point, further increases in input shots did not yield significant performance gains. This suggests a threshold beyond which additional knowledge may not contribute substantially to drug trafficking detection tasks.

\subsection{Limitations and Future Directions}
While our knowledge prompted ChatGPT framework demonstrated impressive performance, several limitations should be considered. First, the framework's effectiveness heavily relies on the quality and comprehensiveness of the domain knowledge integrated into the prompts. Further efforts are required to continuously update and refine the knowledge base to adapt to evolving drug trafficking patterns and techniques.

Second, the evaluation focused on a specific dataset and task. Additional evaluations across different datasets and scenarios are needed to validate the generalizability of the framework.

Furthermore, the interpretability of the framework can be enhanced by exploring techniques such as attention mapping or explainable AI methods. This would enable a better understanding of the model's decision-making process and help address potential biases or limitations.

Future research directions could involve exploring more advanced language models, such as GPT-4 or similar architectures, to improve the framework's performance. Additionally, investigating multi-modal approaches that combine text with other forms of data, such as images or network traffic, could provide a more comprehensive understanding of drug trafficking activities.

\section{Conclusions}\label{sec:6}

This research presented a knowledge-prompted ChatGPT framework for drug trafficking detection. The framework leverages domain-specific knowledge, well-designed prompts, and prompt optimization techniques to improve the performance of the ChatGPT model in accurately identifying and labeling drug trafficking activities. Our framework demonstrates superior performance compared to baseline models, achieving an accuracy of 94.58\%, precision of 96.60\%, recall of 93.42\%, and an F1 score of 94.98\%.
By integrating domain knowledge into the prompts, our framework captures nuanced insights, identifies key indicators, and provides actionable information for the detection of drug trafficking. The knowledge fusion technique, guided by request-related questions and knowledge optimization, contributes to the interpretability and effectiveness of the framework.
We conducted thorough evaluations, including ablation experiments and assessments with varying input shots, to understand the impact of different model components and input variations. The results revealed the importance of prompt design, the value of domain knowledge integration, and the optimal threshold of input shots to maximize performance.

Although our framework exhibits promising results, there are limitations to consider. The framework's effectiveness is heavily dependent on the quality and comprehensiveness of the domain knowledge integrated into the prompts. The generalizability of the framework should be further validated by applying it to diverse datasets and scenarios. Enhancing the interpretability of the framework and exploring advanced language models are potential areas for future research.
The knowledge-prompted ChatGPT framework has great potential in addressing real-world challenges related to drug trafficking detection. By combining the power of language models with domain expertise, the framework contributes to advancing natural language processing techniques for combating illicit activities. Its performance superiority over baseline models shows its practical applicability and potential for implementation in real-world settings.

Overall, this research contributes to the growing body of knowledge-driven approaches in natural language processing and highlights the importance of incorporating domain-specific knowledge to enhance model performance. The framework of knowledge-prompted ChatGPT is a stepping stone for future advancements in the field, with potential applications in various domains beyond drug trafficking detection such as community and key player detection.

\section*{ACKNOWLEDGEMENTS}\label{ACKNOWLEDGEMENTS}

The NSF partially supports this work under grant CMMI-2146076 and CNS-2125958.

 \bibliographystyle{elsarticle-num} 
 \bibliography{cas-refs}





\end{document}